\documentclass{article} 
\usepackage{iclr2026_conference,times}

\usepackage{amsmath,amsfonts,bm}

\def\eqref#1{equation~\ref{#1}}

\def\1{\bm{1}}

\DeclareMathAlphabet{\mathsfit}{\encodingdefault}{\sfdefault}{m}{sl}
\SetMathAlphabet{\mathsfit}{bold}{\encodingdefault}{\sfdefault}{bx}{n}

\usepackage{hyperref}       
\usepackage{url}            
\usepackage{booktabs}       
\usepackage{amsfonts}       
\usepackage{nicefrac}       
\usepackage{xcolor}         
\usepackage{amsmath}
\usepackage{xspace}
\usepackage{amssymb}

\usepackage{algorithm}
\usepackage{multirow}

\usepackage{algorithmic}

\usepackage[most]{tcolorbox}
\usepackage{tabularx}
\usepackage{colortbl}
\usepackage{enumitem}
\usepackage{subcaption}

\newcommand{\ourmodel}{MISP-DPO\xspace}

\newtheorem{lemma}{Lemma}[section]

\title{Importance Sampling for Multi-Negative Multimodal Direct Preference Optimization}

\author{Xintong Li$^{1}$\thanks{These authors contributed equally to this work.}, Chuhan Wang$^{1*}$, Junda Wu$^1$, Rohan Surana$^1$, Tong Yu$^2$, \\
\textbf{Julian McAuley$^1$ , Jingbo Shang$^1$} \\
$^1$University of California, San Diego \quad 
$^2$Adobe Research \\
\texttt{\{xil240,chw136,juw069,rsurana,jmcauley,jshang\}@ucsd.edu} \\
\texttt{tyu@adobe.com} \\
}

\iclrfinalcopy 
\begin{document}

\maketitle

\begin{abstract}

Direct Preference Optimization (DPO) has recently been extended from text-only models to vision-language models. 
However, existing methods rely on oversimplified pairwise comparisons, generating a single negative image via basic perturbations or similarity-based retrieval, which fail to capture the complex nature of multimodal preferences, inducing optimization bias and hallucinations.
To address this issue, we propose \ourmodel, the first framework to incorporate \emph{multiple}, semantically \emph{diverse} negative images in multimodal DPO via the Plackett-Luce model. 
Our method embeds prompts and candidate images in CLIP (Contrastive Language–Image Pre-training) space and applies a sparse autoencoder to uncover semantic deviations into interpretable factors.
Negative samples are selected based on reconstruction difficulty, semantic deviation from the positive, and mutual diversity, yielding broader and more informative supervision.
To handle multi-negative comparisons, we adopt a Plackett–Luce objective and introduce an importance sampling strategy that improves training efficiency.
Experiments across five diverse benchmarks demonstrate that \ourmodel consistently improves multimodal alignment over prior methods, validating the effectiveness of semantic-aware, multi-negative sampling in preference-based learning.

\end{abstract}
\section{Introduction}
\begin{figure*}[t]
    \centering
    \includegraphics[page=1, clip, trim=0 370 0 50, width=\linewidth]{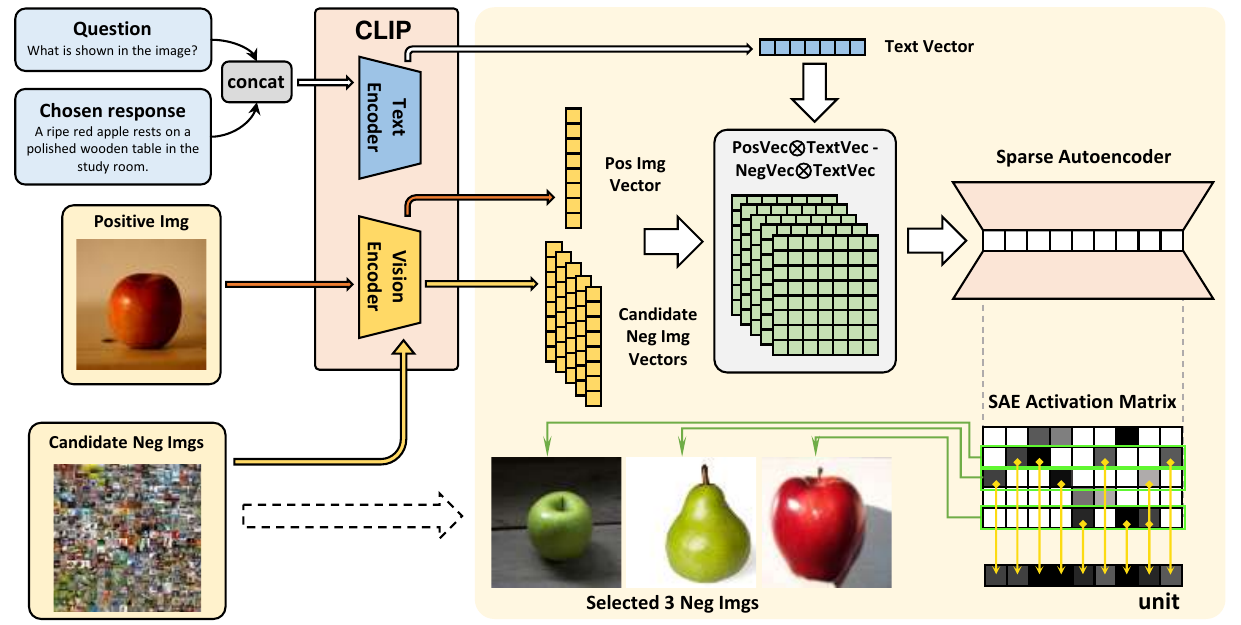}
    \caption{Overview of the MISP-DPO framework, which integrates CLIP encoding and sparse autoencoder–guided selection to identify diverse negatives for multi-negative preference optimization.}
    \label{fig:overview}
    \vspace{-3mm}
\end{figure*}

Direct Preference Optimization (DPO)~\citep{rafailov2023direct, amini2024direct} has shown great promise for aligning language models by learning from pairwise comparisons, bypassing the need for explicit reward modeling. 
Recent efforts have extended DPO to multimodal contexts, enhancing vision-language model (VLM) alignment through image-text feedback\citep{wang2024mdpo, jiang2024modality, deng2024enhancing, fu2025chip, liu2025adpo, wu2025symmetrical, xing2025re}. 
However, simply extending textual preference data to multimodal scenarios often introduces new challenges, particularly exacerbating hallucinations~\citep{wang2024mdpo, fu2025chip, wu2025symmetrical}.
Existing multimodal DPO methods generate only a single negative image per comparison, typically via adversarial cropping, random perturbations, or similarity-based retrieval~\citep{liu2025adpo, fu2025chip, wu2025symmetrical, xing2025re}.
This oversimplifies the rich space of visual negatives, reducing supervision to a single dimension and limiting the model's ability to generalize.
For instance, as illustrated in Figure~\ref{fig:overview}, avoiding a single negative depicting a ``green apple'' might teach the model to reject green hues but ignore mismatched contexts like ``kitchen counter'' or incorrect objects like ``pear.''
By optimizing against narrow, one-dimensional deviations, models risk spurious correlations, bias amplification, and persistent hallucinations.

The core challenge is that images lack explicit, compositional units like text tokens, making it difficult to isolate meaningful visual deviations~\citep{sahin2024enhancing, zeng2024investigating, zheng2024iterated, hsieh2023sugarcrepe, kamath2024hard, kamath2023text}. 
Naive perturbations often destroy overall coherence without isolating meaningful deviations, making it difficult to systematically explore the negative factors of model weaknesses.
Effective learning requires disentangling and surfacing multiple latent error factors while maintaining prompt relevance.
Existing methods incorporate these factors into a single negative example, leaving models blind to orthogonal error types.

To address this, we propose \ourmodel, 
the first framework to introduce \emph{multi}-negative, semantically \emph{diverse} supervision into multimodal DPO. 
Our approach consists of two stages, 
\begin{itemize}[nosep,leftmargin=*]
    \item
        In the first stage, we select diverse image-side negatives from a large open-domain pool. 
        Prompts and candidate images are embedded in CLIP~\citep{radford2021learning} space, and a sparse autoencoder (SAE) decomposes their semantic differences into disentangled latent factors (e.g., object, color, layout). 
        We prioritize negatives based on: (1) reconstruction error (informativeness), (2) semantic deviation from the positive sample, and (3) mutual diversity, ensuring broad coverage of negative types.
    \item 
        In the second stage, we integrate these multiple negatives into a generalized DPO objective using the Plackett–Luce model. 
        Rather than relying on binary comparisons, our approach ranks a positive image above a diverse set of negatives, forcing the model to resolve multiple constraints simultaneously. 
        We further introduce an importance sampling scheme guided by SAE-derived scores, improving training efficiency.
\end{itemize}
We evaluate \ourmodel on five multimodal benchmarks~\citep{sun2023aligning, guan2024hallusionbench, lu2024wildvision, tong2024eyes, li2023seed} focused on hallucination reduction and visual grounding. Our method consistently outperforms strong baselines, achieving notable hallucination reduction and improved alignment, including a 30.09\% average improvement over LLaVA-v1.5-7B~\citep{liu2023visual}.

Our contributions are as follows:
\begin{itemize}[nosep,leftmargin=*]
\item We propose the first framework to incorporate multi-negative supervision into multimodal DPO, leveraging semantic diversity to systematically reduce hallucinations.
\item We introduce an efficient negative sampling method based on CLIP embeddings and SAE–guided importance sampling, providing semantically informative negative examples.
\item Extensive evaluations demonstrate that our method substantially reduces hallucinations and achieves robust multimodal alignment across multiple benchmarks.
\end{itemize}

\section{Related Works} 
\textbf{Multimodal Direct Preference Optimization.}   
DPO~\citep{rafailov2023direct} has become a widely adopted method for aligning LLMs with human preferences due to its simplicity and stability. However, when extended to multimodal scenarios, especially for hallucination-prone tasks, standard DPO often fails to effectively incorporate visual signals, leading models to overfit textual biases and ignore image-grounded constraints.
To mitigate this, recent works have adapted DPO for multimodal hallucination reduction by incorporating visual preference supervision. mDPO~\citep{wang2024mdpo} introduces conditional preference learning and reward anchoring, using lightweight perturbations (e.g., cropping, diffusion) to construct visual negatives. CHiP~\citep{fu2025chip} further complements this with hierarchical textual supervision and a visual contrastive loss to better align fine-grained text and image semantics.
While both methods demonstrate notable gains on hallucination benchmarks, they rely on limited forms of visual augmentation, often constrained to local perturbations with narrow semantic variation. Other approaches, such as S-VCO~\citep{wu2025symmetrical} and Re-Align~\citep{xing2025re}, explore counterfactual or retrieval-based visual negative generation, but at the cost of high computation and limited scalability. 
In this work, we follow the hallucination-centric preference optimization paradigm initiated by mDPO, and propose a scalable framework for generating informative visual negatives tailored for multimodal preference learning. 


\textbf{Multi-negative Preference Optimization.}  
Recent works in textual and recommendation domains~\citep{amini2024direct, shi2024direct, baruchpreference} have extended DPO to multi-negative settings, ranking positives above multiple negatives to enhance robustness. For example, Softmax-DPO~\citep{chen2024softmax} and DMPO~\citep{shi2024direct} adopt soft ranking or Plackett–Luce objectives to reduce noise sensitivity. However, such techniques remain underexplored in vision-language models, where negatives must capture subtle cross-modal semantic shifts.
Inspired by findings in attribute-based recognition~\citep{yan2023learning, wang2024disentangled} showing that compact, curated subsets can match large noisy sets, we adapt this insight to multimodal preference learning. Our framework uses a sparse autoencoder in CLIP space to select semantically diverse negatives, enabling importance-weighted ranking over multiple contrastive examples and capturing fine-grained failure modes more effectively.
\section{Preliminaries} 

\subsection{Multimodal Direct Preference Optimization}

DPO~\citep{rafailov2023direct} provides a principled way to align a learned policy with human preference judgments without explicitly modeling rewards.  In the RLHF framework, solving for the optimal policy $\pi^*$ under a fixed reference policy $\pi_{\mathrm{ref}}$ yields a latent reward function
\begin{equation}
\label{eq:dpo-policy}
r(x,y) 
= \beta \log\frac{\pi^*(y\mid x)}{\pi_{\mathrm{ref}}(y\mid x)} + Z(x),
\end{equation}
where $\beta$ scales the strength of alignment and $Z(x)$ is a prompt‐dependent normalizer.  Substituting this into the Bradley–Terry–Luce model and dropping $Z(x)$ gives a simple training objective for a parametric policy $\pi_\theta$,
\begin{equation}
\label{eq:dpo-loss}
\mathcal{L}_{\mathrm{DPO}}(\theta)
= -\mathbb{E}_{(x,y_p,y_n)\sim D}
    \Bigl[\log \sigma\bigl(\beta\log\tfrac{\pi_\theta(y_p\mid x)}{\pi_{\mathrm{ref}}(y_p\mid x)}
    \;-\;\beta\log\tfrac{\pi_\theta(y_n\mid x)}{\pi_{\mathrm{ref}}(y_n\mid x)}\bigr)\Bigr].
\end{equation}


Recent work extends DPO to vision-language models (VLMs) by incorporating visual preferences. Let $x$ denote the multimodal prompt, $m_p$ a preferred image aligned with textual response $y_p$, and $m_n$ a rejected image. The multimodal reward $r(m,x,y)$ now depends on visual grounding, with preferences modeled as,
\begin{equation*}
\label{eq:dpo-pref-m}
p^{*}(y_{p}\succ y_{n}\mid m,x)
=\sigma\bigl(r(m,x,y_{p})-r(m,x,y_{n})\bigr).
\end{equation*}
To ensure a fair comparison across images, we hold $y_p$ fixed and vary only the image input, the multimodal DPO loss~\citep{wang2024mdpo, fu2025chip, wu2025symmetrical} focuses on visual discrimination,
\begin{equation}
\label{eq:dpo-loss-m}
\mathcal{L}_{\mathrm{DPO}_{\mathrm{img}}}(\theta)
=-\mathbb{E}_{(m_p, m_n,x,y_{p})\sim D}
\Bigl[\log\sigma\!\bigl(\beta\log\tfrac{\pi_{\theta}(y_{p}\mid m_p,x)}{\pi_{\mathrm{ref}}(y_{p}\mid m_p,x)}
-\beta\log\tfrac{\pi_{\theta}(y_p\mid m_n,x)}{\pi_{\mathrm{ref}}(y_p\mid m_n,x)}\bigr)\Bigr].
\end{equation}

This formulation supports joint optimization over visual and textual inputs, enabling the policy to associate preferred images with relevant multimodal features.  

\subsection{Multi‐negative Preference Optimization}
Multi-negative preference optimization~\citep{chen2024softmax} extends the Direct Preference Optimization approach~\citep{rafailov2023direct}, enabling language models to be trained against several negative preferences rather than just one. Instead of using the Bradley–Terry formulation for single pairwise comparisons, this method adopts the Plackett–Luce model~\citep{plackett1975analysis,luce1959individual} to score a target choice in relation to an entire set of inferior alternatives.

Given a prompt \(x\), a preferred response \(y_p\), and a set of \(N\) non‐preferred responses \(\mathcal{Y}_n = \{y_n^i\}_{i=1}^N\), the Plackett–Luce probability that \(y_p\) is ranked above all \(y_n^i\) is
\begin{equation}
\label{eq:pl-latent}
p^*(y_p \succ \mathcal{Y}_n \mid x)
= \frac{\exp\bigl(r(x,y_p)\bigr)}
       {\exp\bigl(r(x,y_p)\bigr)\;+\;\sum_{i=1}^N \exp\bigl(r(x,y_n^i)\bigr)},
\end{equation}
where \(r(x,y)\) is the latent reward function.  Substituting  
\[
r(x,y)
= \beta\,\log\frac{\pi_\theta(y\mid x)}{\pi_{\mathrm{ref}}(y\mid x)} \;+\; Z(x)
\]
and noting that \(Z(x)\) cancels in the ratio gives
\[
p^*(y_p \succ \mathcal{Y}_n \mid x)
= \frac{1}
       {1 + \sum_{i=1}^N \exp\bigl(\beta\,\Delta_i\bigr)},
\quad
\Delta_i
= \log\frac{\pi_\theta(y_n^i\mid x)}{\pi_{\mathrm{ref}}(y_n^i\mid x)}
- \log\frac{\pi_\theta(y_p\mid x)}{\pi_{\mathrm{ref}}(y_p\mid x)}.
\]
Hence the multi‐negative DPO training objective becomes
\begin{equation}
\label{eq:mn-dpo-loss}
\mathcal{L}_{\mathrm{MN\text{-}DPO}}(\theta)
= -\,\mathbb{E}_{(x,y_p,\mathcal{Y}_n)\sim D}
    \Bigl[\log \sigma\bigl(-\,\log\sum_{i=1}^N 
      \exp\bigl(\beta\,\Delta_i\bigr)\bigr)\Bigr].
\end{equation}
Notably, when \(N=1\), \(\mathcal{L}_{\mathrm{MN\text{-}DPO}}\) in \eqref{eq:mn-dpo-loss} reduces exactly to the single‐negative DPO loss.

\section{Framework}
We propose \ourmodel, a framework that address the limitations of single-negative supervision by introducing multi-negative learning through two core components: (1) a diverse negative sampling strategy using sparse autoencoders to identify semantically meaningful deviations, and (2) a generalized Plackett-Luce ranking objective that integrates multiple negatives to promote robust alignment. An overview of the framework is shown in Figure~\ref{fig:overview}.

\subsection{Multi-negative Objectives} \label{sec:setting}

Due to the limitations of single-negative supervision and the inherently multi-faceted nature of visual errors, we extend multimodal DPO to a multi-negative preference optimization setting.
Let $\pi_\theta$ denote the VLM policy to be optimized. Each training instance consists of a multimodal prompt $x$, a preferred image $m_p$ paired with an aligned textual response $y_p$, and a set of $N$ negative images $S_n = \{m_n^i\}_{i=1}^N$ from open-domain sources. 
Following the Plackett-Luce formulation from Eq.~\eqref{eq:pl-latent}, we adapt \eqref{eq:mn-dpo-loss} to visual preferences,
\begin{equation}
\label{eq:sdpo-image-loss}
\mathcal{L}_{\mathrm{img}}(\theta;S_n) = \log \sigma\left( -\log \sum_{i \in S_n} \exp\left( \beta \log\frac{\pi_\theta(y_p \mid x, m_n^i)}{\pi_{\mathrm{ref}}(y_p \mid x, m_n^i)} - \beta \log\frac{\pi_\theta(y_p \mid x, m_p)}{\pi_{\mathrm{ref}}(y_p \mid x, m_p)} \right) \right)
\end{equation}
This extends Eq.~\eqref{eq:dpo-loss-m} to multiple negatives through the softmax aggregation and encourages the model to assign higher preference scores to the correct image $m_p$ compared to all negative images in $S_n$, thereby promoting more robust visual grounding.
\begin{lemma}[Gradient Decomposition]\label{lemma:grad-decomp}
Defining the preference advantage of each negative image and the preference distribution as
\begin{equation*}
a_i = \beta \left(\log\frac{\pi_\theta(y_p \mid x, m_n^i)}{\pi_{\mathrm{ref}}(y_p \mid x, m_n^i)} - \log\frac{\pi_\theta(y_p \mid x, m_p)}{\pi_{\mathrm{ref}}(y_p \mid x, m_p)}\right) , \quad p_\theta(m_n^i \mid x,m_p,y_p)
=\frac{\exp(a_i)}{\sum_{j=1}^N \exp(a_j)}.
\end{equation*}
Then the gradient of \eqref{eq:sdpo-image-loss} decomposes as,
\begin{align}
\label{eq:mn-dpo-grad}
\nabla_\theta \mathcal{L}_{\mathrm{img}}(\theta;\mathcal{S}_n)
&=\beta\sigma\Bigl(\log\sum_{i=1}^N \exp(a_i)\Bigr)
\sum_{i=1}^N p_\theta(m_n^i\mid x,m_p,y_p)
\Delta_\theta(m_n^i,m_p\mid x,y_p),
\end{align}
where $\Delta_\theta(m_n^i,m_p\mid x,y_p)
=\nabla_\theta \log\pi_\theta(y_p\mid x,m_n^i)
-\nabla_\theta \log\pi_\theta(y_p\mid x,m_p)$.
\end{lemma}

This result shows that the gradient is a weighted combination of correction signals across the image space, offering interpretability in terms of how the model adjusts its predictions in response to each visual discrepancy.

Although Eq.(~\ref{eq:sdpo-image-loss}) and its gradient give an unbiased update, they require drawing a large set of negatives from the true $p_\theta(m_n^i \mid x, m_p, y_p)$ and computing $S(\{m_n^i\})$. 
In realistic image domains, neither step is tractable. To alleviate this, we introduce a learnable distribution $q_\phi(m_n\mid x,m_p,y_p)$ to sample a small candidate pool $\tilde S_n$. 
Rewriting the gradient illustrated in Lemma~\ref{lemma:grad-decomp} as an expectation under $q_\phi$ gives a \emph{importance‐sampling} estimator,
\begin{equation}
\label{eq:is-grad}
\nabla_\theta \mathcal{L}_{\mathrm{img}}(\theta;\tilde{\mathcal{S}}_n)
=
\beta\sigma\Bigl(\log\sum_{i\in\tilde{\mathcal{S}}_n}\exp(a_i)\Bigr)
\sum_{i\in\tilde{\mathcal{S}}_n}
\frac{\exp(a_i)}{q_\phi(m_n^i\mid x,m_p,y_p)}
\Delta_\theta(m_n^i,m_p\mid x,y_p).
\end{equation}

To encourage joint reasoning across modalities, we extend our framework by incorporating textual preference supervision. We follow recent multimodal DPO methods and replace traditional text-only preferences with image-grounded negative responses $y_n$ for the same prompt and image $m_p$. The corresponding DPO loss is,
\begin{align}
\label{eq:text-dpo}
\mathcal{L}_{\mathrm{text}}(\theta;\tilde{\mathcal{S}}_n)
&=\log\sigma\Bigl(\beta\log\frac{\pi_\theta(y_p\mid x,m_p)}{\pi_{\mathrm{ref}}(y_p\mid x,m_p)}
-\beta\log\frac{\pi_\theta(y_n\mid x,m_p)}{\pi_{\mathrm{ref}}(y_n\mid x,m_p)}\Bigr).
\end{align}
Our final loss combines both visual and textual preference signals,
\begin{equation}
\label{eq:total-loss}
\mathcal{L}(\theta; \tilde{\mathcal{S}}_n)
=\mathcal{L}_{\mathrm{img}}(\theta;\tilde{\mathcal{S}}_n)
+\lambda\,\mathcal{L}_{\mathrm{text}}(\theta; \tilde{\mathcal{S}}_n).
\end{equation}
where $\lambda$ balances the contributions of image-based and text-based supervision. This unified formulation supports joint alignment across modalities, improving robustness and alignment quality in VLMs.

\subsection{Importance Sampling via Sparse Autoencoder}
To address the limitations of existing methods that rely on simplistic, one-dimensional negatives, we employ SAEs to disentangle and surface semantically meaningful variations in the visual space. By providing a structured and interpretable latent representation, SAEs enable principled importance sampling over diverse negative examples—prioritizing those that capture distinct failure modes and are most informative for effective preference learning.

\textbf{Embedding and Difference Vectors.}  
Let $\mathcal{T}=\{(m_p,x)\}$ be the training set of positive image–prompt pairs. We use CLIP’s image and text encoders, $f_v$ and $f_t$, to obtain $d$-dimensional embeddings $h_v = f_v(m_p)$ and $h_t = f_t(x)$, then fuse them via outer product and vectorization $e = \mathrm{vec}\bigl(h_v\times h_t^\top\bigr)\in\mathbb{R}^{d^2}.$
For each negative candidate $m_n^i$, we form the difference vector
$$
d_i \;=\; e(m_p,x)\;-\;e(m_n^i,x).
$$
\textbf{Sparse Autoencoder Training.}  
We train an SAE with encoder $\mathcal{E}$ and decoder $\mathcal{D}$ to decompose $d_i$ into sparse latent factors. The loss combines reconstruction fidelity and activation sparsity,
\begin{equation}
\label{eq:sae-loss}
\mathcal{L}_{\mathrm{SAE}}
=\frac{1}{|\mathcal{T}|N}
\sum_{(m_p,x)\in\mathcal{T}}
\sum_{i=1}^{N}
\bigl\lVert d_i - \mathcal{D}\bigl(\mathcal{E}(d_i)\bigr)\bigr\rVert_2^2
+
\gamma
\sum_{j=1}^{H}
\mathrm{KL}\bigl(\rho \big\| \hat\rho_j\bigr),
\end{equation}
where$\hat\rho_j$ is the average activation of hidden unit $j$, $\rho\in(0,1)$ is the target average activation, and $\gamma$ balances reconstruction against sparsity.

\textbf{Diverse Negative Selection.}  
We score each candidate $m_n^i$ by,
\begin{equation}
    s_i = \frac{\bigl\lVert d_i - \mathcal{D}(\mathcal{E}(d_i))\bigr\rVert_2^2}{\max_j \ell_j}+ \,\frac{\bigl\lVert \mathcal{E}(d_i)\bigr\rVert_1}{\max_j v_j},
\end{equation}
where $\ell_j$ and $v_j$ are the reconstruction error and activation magnitude across all candidates.
To choose the final top-$K$ negatives $\tilde{\mathcal{S}}_n$, we run a greedy selection that maximizes coverage of distinct error types while emphasizing hard negatives. We illustrate this algorithm in detail in Algorithm~\ref{alg:greedy-topk}.

\begin{algorithm}[t]
\caption{Greedy Diversity-Promoting Selection}\label{alg:greedy-topk}
\begin{algorithmic}[1]
\STATE \textbf{Input:} Difference vectors $\{d_i\}_{i=1}^N$, scores $\{s_i\}$, encoder $\mathcal{E}$, selection size $K$
\STATE $\tilde{\mathcal{S}}_n \leftarrow \emptyset$
\WHILE{$|\tilde{\mathcal{S}}_n| < K$}
  \STATE $i^* \leftarrow \arg\max_{i\notin\tilde{\mathcal{S}}_n}\Bigl(s_i + \beta\,\min_{j\in\tilde{\mathcal{S}}_n}(1-\mathrm{cos}(\mathcal{E}(d_i),\mathcal{E}(d_j)))\Bigr)$
  \STATE $\tilde{\mathcal{S}}_n \leftarrow \tilde{\mathcal{S}}_n \,\cup\,\{i^*\}$
\ENDWHILE
\STATE \textbf{Output:} selected negatives set $\tilde{\mathcal{S}}_n$
\end{algorithmic}
\end{algorithm}

The selected set $\tilde{\mathcal{S}}_n$ is then used in the importance‐sampling gradient estimator of Eq.~\eqref{eq:is-grad}.

\section{Experiment}

\subsection{Experimental Settings}
\textbf{Models.} We apply \ourmodel to three widely-used multimodal LLMs: LLaVA-1.5-7B-HF, Qwen2.5-VL-7B, and Qwen2.5-VL-3B. These models are chosen due to their open availability, competitive performance, and diverse architectural designs~\citep{chen2024expanding, zhang2024mm}. LLaVA-1.5-7B-HF~\citep{liu2023visual} integrates CLIP as the vision encoder with Vicuna-1.5-7B as the language backbone. Qwen2.5-VL-7B~\citep{bai2025qwen2} uses a proprietary vision module and the strong Qwen2.5-7B language model. Qwen2.5-VL-3B is a lightweight 3B variant of the same architecture, providing a better balance between efficiency and capability. 

\textbf{Training data.} We choose RLHF-V-Dataset~\citep{yu2024rlhf} as our training dataset. It contains more than 5K samples, each with an image and a pair of text responses indicating preference. RLHF-V provides fine-grained, segment-level human feedback on diverse vision-language instructions, which has been shown to largely reduce hallucination while preserving informativeness. We treat the paired image as the positive sample and select 3 negative images per sample from COCO training dataset~\citep{lin2014microsoft} using our importance sampling method, enabling effective training in our \ourmodel framework.

\textbf{Baselines.} We compare \ourmodel against five baselines: (1) the pretrained model without preference tuning, (2) standard DPO~\citep{rafailov2023direct}, which uses only a single text preference without any image-based supervision, (3) mDPO~\citep{wang2024mdpo} and (4) CHiP~\citep{fu2025chip}, both of which incorporate image preferences but rely on a single negative image per comparison, and (5) a variant of our framework that uses multi-negative image preference optimization with negatives randomly sampled from the COCO dataset. 
All methods are trained under the same settings for a fair comparison.

\textbf{Evaluation Benchmarks.} 
We evaluate \ourmodel and all baselines across five benchmarks spanning hallucination detection and vision-centric reasoning.
MMHal-Bench\citep{sun2023aligning} is a hallucination-focused VQA benchmark covering 8 question types and 12 object categories. 
HallusionBench\citep{guan2024hallusionbench} measures visual and factual hallucinations; we report all-answer accuracy (aA), figure-based accuracy (fA), and question-type accuracy (qA).
POPE\citep{li2023evaluating} evaluates object hallucination in VLMs via Yes/No probing under random, popular, and adversarial object settings.
WildVision\citep{lu2024wildvision} evaluates real-world user preference alignment with 500 curated human-model interaction samples; we report reward score and win rate. 
MMVP\citep{tong2024eyes} assesses fine-grained visual reasoning using CLIP-blind image pairs, with accuracy reported over 135 zero-shot questions across 9 pattern types. 
Except for MMHal-Bench, all evaluations are conducted using VLMEvalKit\citep{duan2024vlmevalkit}, an open-source evaluation toolkit for vision-language models. For MMHal-Bench, we use GPT-4.1-mini\citep{achiam2023gpt} as the evaluator and report overall response quality and hallucination rate.

\textbf{Implementation Details.} 
For training the Sparse Autoencoder, we set the latent dimension to 128 and the sparsity weight $\gamma$ to 1, balancing reconstruction fidelity with latent sparsity.
Following prior work on multi-negative preference optimization~\citep{chen2024softmax}, we select three negative images per instance.
For multimodal DPO training, we set the supervision balance parameter $\lambda$ to 1 to equally weight image-based and text-based preferences. 
All models are fine-tuned using 2 NVIDIA A100 GPUs, with a per-device batch size of 2, gradient accumulation steps of 8 (yielding an effective batch size of 32), and a learning rate of $10^{-5}$. 
The preference optimization coefficient $\beta$ is set to 0.5. Following prior work on mDPO, we adopt LoRA for parameter-efficient tuning, with a rank of 64 and scaling factor $\alpha = 128$. 
For baseline methods, we strictly follow their original settings: mDPO is trained for 3 epochs with the same learning rate (1e-5), $\beta = 0.1$; CHiP is trained for 3 epochs with a batch size of 32, a learning rate of 5e-7, $\beta = 0.5$, and full-parameter finetuning. 

\begin{table*}[tbp]
  \centering
  \renewcommand{\arraystretch}{1.2}
  \resizebox{\linewidth}{!}{
    \setlength{\tabcolsep}{4pt}
    \begin{tabular}{clccccccclc|c}
      \toprule
      & & 
      \multicolumn{6}{c}{\textbf{Hallucination}} & 
      \multicolumn{3}{c}{\textbf{Vision-Centric}} & 
      \textbf{Total}
       \\
      \cmidrule(r){3-8} \cmidrule(r){9-11} \cmidrule(r){12-12}
      & \multirow{2}{*}{\textbf{Benchmarks}} 
        & \multicolumn{2}{c}{\textbf{MMHalBench}} 
        & \multicolumn{3}{c}{\textbf{HallusionBench}} 
        & \textbf{POPE} 
        & \multicolumn{2}{c}{\textbf{WildVision}} 
        & \textbf{MMVP} 
        & \textbf{avg\_impr.} \\
      & 
        & \textbf{Score} ($\uparrow$) & \textbf{HalRate} ($\downarrow$) 
        & \textbf{aA} ($\uparrow$) & \textbf{fA} ($\uparrow$) & \textbf{qA} ($\uparrow$) 
        & \textbf{Acc.} ($\uparrow$) 
        & \textbf{Reward} ($\uparrow$) & \textbf{WinRate} ($\uparrow$) 
        & \textbf{Acc.} ($\uparrow$)
        & \textbf{over BASE} \\
      \midrule
    \multirow{6}{*}{\rotatebox{90}{\textbf{llava-1.5-7b-hf}}} 
    & Base          & 2.78 & 51.04 & 47.73 & 17.63 & 12.30 & \textbf{84.37} & -55.7 & 17.0 & 60.67 & 0\% \\
    & DPO           & 3.29 & 37.50 & 55.62 & 22.83 & 22.63 & 83.02 & -52.7 & 18.4 & 62.66 & +21.13\% \\
    & mDPO          & 2.99 & 49.81 & 47.32 & 20.52 & 13.19 & 83.25 & -62.1 & 14.6 & 58.33 & +0.22\%  \\
    & CHiP          & 3.13   & 34.04  & 51.95 & 17.92 & 19.78 & 82.56 & -68.4  & 12.2  & 52.33 & +5.59\%  \\
    & Random        & 3.42 & 36.46 & 55.94 & 23.69  & 22.63 & 82.61 & -51.3 & 18.3 & 60.33 &  +22.23\% \\
    & \textbf{\ourmodel}     & \textbf{3.51}  & \textbf{32.29} & \textbf{57.52}   & \textbf{25.43}   & \textbf{24.83} & 83.94 & \textbf{-46.4} & \textbf{20.6} & \textbf{63.00} & \textbf{+30.09\% } \\
    \midrule
    \multirow{6}{*}{\rotatebox{90}{\textbf{Qwen2.5-VL-7B}}} 
    & Base          & 4.61  & 18.09 & 70.45 & 43.06 & 45.27 & 87.65 & \textbf{33.5} & \textbf{69.2} & 77.67 & 0\% \\
    & DPO           & 4.92  & \textbf{11.46} & 69.92  & 42.48 & 43.51 & 87.46 & 32.8 & 68.6 & 78.00 & +3.85\%  \\
    & mDPO          & 5.01  & 14.89 & 67.40 & 41.33 & 42.20 & 87.02 & 28.5  & 66.2  & 76.33 & -1.16\% \\
    & CHiP          & 5.02  & 13.83 & 66.14  & 39.02  & 40.88 & 88.18 & 28.5  & 66.4  & 77.00  & -1.33\% \\
    & Random        & 4.75  & 16.67 & 70.24 & 42.48 & 43.95 & 87.60  & 30.7 & 67.8 & 78.33 & -0.36\% \\
    & \textbf{\ourmodel}     & \textbf{5.05} & \textbf{11.46} & \textbf{71.24} & \textbf{43.77} & \textbf{45.61}  & \textbf{88.66}  & 32.4 & 68.4 & \textbf{79.00} & \textbf{+5.35\%} \\
    \midrule
    \multirow{6}{*}{\rotatebox{90}{\textbf{Qwen2.5-VL-3B}}} 
    & Base          & 4.20  & 22.34 & 64.67 & 37.57 & 36.70 & 87.48 & -0.1 & 46.6 & 70.60 & 0\% \\
    & DPO           & 4.50  & 18.75 & 65.19  & 36.41 & 37.14 & 87.42 & 7.5  & 51.2 & 71.33   & +13.12\% \\
    & mDPO          & 4.47  & 21.28 & 62.88 & 35.84 & 37.14 & 87.65 & 7.2  & 50.8  & 69.33 & +10.51\% \\
    & CHiP          & 4.51  & 15.96 & 62.14  & 36.13 & 34.29 & 87.30 & 6.3 & 51.0  & 70.33 & +11.81\% \\
    & Random        & 4.27  & 16.67 & 64.98   & \textbf{38.44}  & 37.36 & 87.52    & 3.6   & 48.6 & 74.25  & +8.57\% \\
    & \textbf{\ourmodel}     & \textbf{4.61}  & \textbf{13.54} & \textbf{65.51} & \textbf{38.44}  & \textbf{38.02} & \textbf{87.77} & \textbf{8.6} & \textbf{52.4} & \textbf{72.00}  & \textbf{+19.89\%} \\
    \bottomrule
    \end{tabular}
  }
  \caption{
Comparison of MISP-DPO against baseline methods across five vision–language benchmarks and three model backbones. The benchmarks cover hallucination detection and vision-centric reasoning. Average improvement over BASE is reported. $\uparrow$: higher is better; $\downarrow$: lower is better.
}
\vspace{-4mm}
\label{tab:main_table}
\end{table*}

\subsection{Overall Performance Improvement}
Table~\ref{tab:main_table} shows the performance of \ourmodel and baselines across five representative benchmarks, grouped into two categories: hallucination detection (MMHalBench, HallusionBench, POPE) and vision-centric reasoning (WildVision, MMVP). Our proposed \ourmodel consistently achieves superior results over all evaluation domains and model backbones.

The largest gains appear on hallucination benchmarks. \ourmodel substantially reduces hallucination rates on MMHalBench (e.g., 32.29\%, 11.46\%, and 13.54\% across different backbones) while also achieving the highest accuracy across all HallusionBench metrics. 
POPE further confirms these advantages. These improvements stem from the combination of diverse negative sampling, which exposes the model to varied error types such as object mismatches and attribute distortions, and importance sampling, which prioritizes hard negatives with high reconstruction errors from SAE, leading to stronger visual grounding.
On vision-centric reasoning tasks, \ourmodel also provides consistent gains. For example, it achieves the best reward (+8.6) and win rate (52.4) on WildVision for Qwen2.5-VL-3B, while also outperforming baselines on MMVP across different model sizes. These results suggest that our method not only suppresses hallucinations but also enhances the model’s ability to generate fine-grained, visually aligned responses.


We also conduct experiments under different $\beta$ values on MMHalBench to balance reward learning and regularization. Figure~\ref{fig:beta_value} reveals performance peaks at $\beta$ ranging from 0.45 to 0.75, with degradation at extremes $\beta = 0.1/1.0$. We choose $\beta = 0.5$ for optimal trade-off between hallucination control and response quality, as it maximizes accuracy while minimizing hallucination rates across all backbones.

\begin{figure}[t]
  \centering
  \begin{subfigure}[t]{0.32\linewidth}
    \centering
    \includegraphics[width=\linewidth]{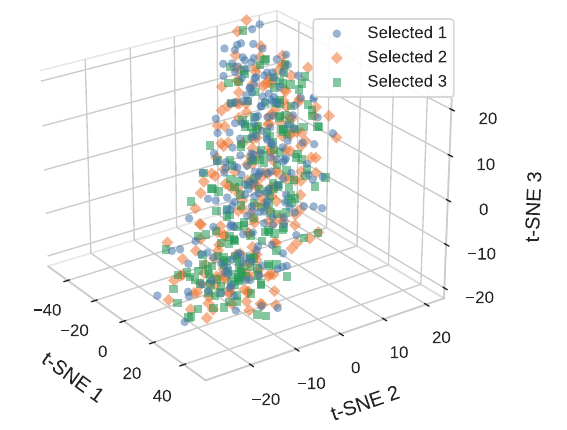}
  \end{subfigure}\hfill
  \begin{subfigure}[t]{0.285\linewidth}
    \centering
    \includegraphics[width=\linewidth]{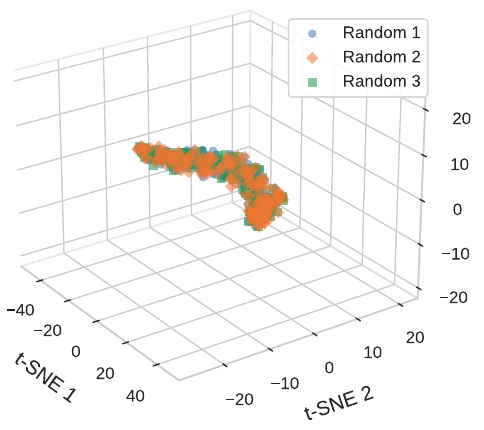}
  \end{subfigure}\hfill
  \begin{subfigure}[t]{0.32\linewidth}
    \centering
    \includegraphics[width=\linewidth]{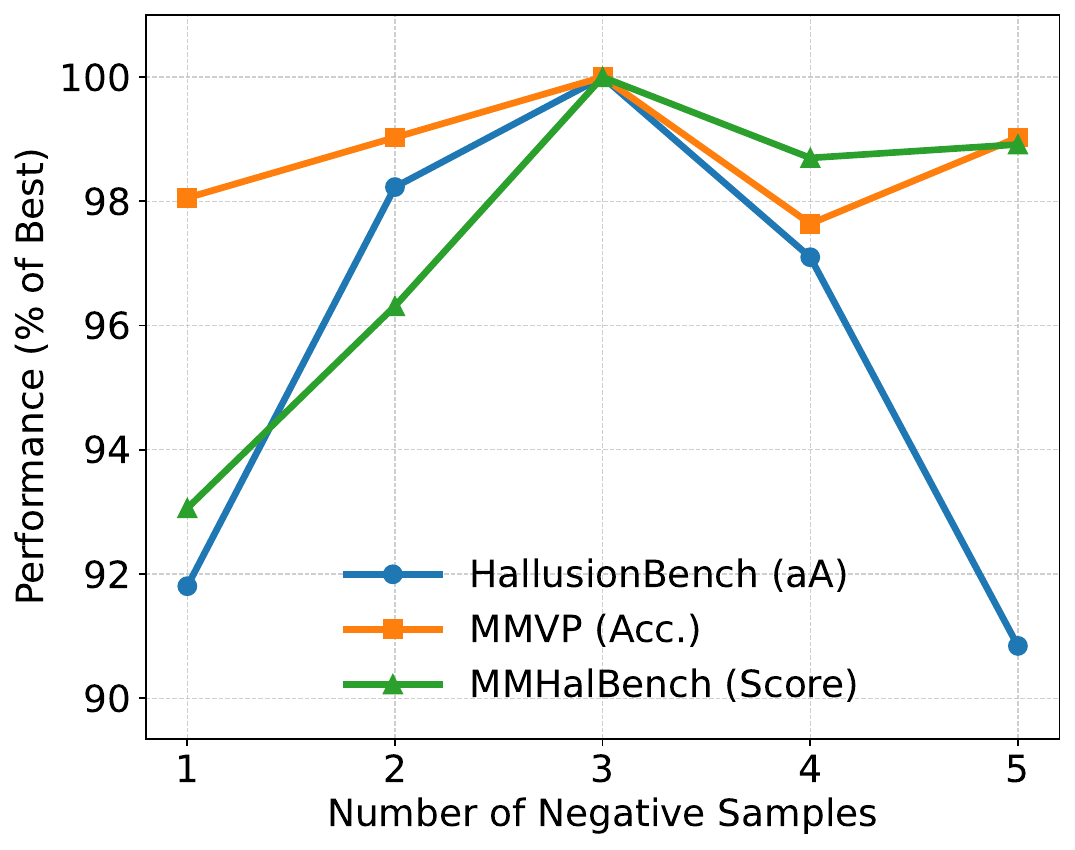}
  \end{subfigure}
  \vspace{-2mm}
  \caption{t-SNE visualizations and benchmark results for negative sampling.
  \textbf{Left:} importance-sampled negatives selected using SAE scoring exhibit broad semantic dispersion across three selections.
  \textbf{Middle:} randomly sampled negatives form tight, low-diversity clusters.
  \textbf{Right:} performance across benchmarks with different numbers of negatives selected by \ourmodel.}
  \label{fig:tsne-sampling}
  \vspace{-2mm}
\end{figure}

\begin{figure}[t]
    \centering
    \includegraphics[width=\linewidth]{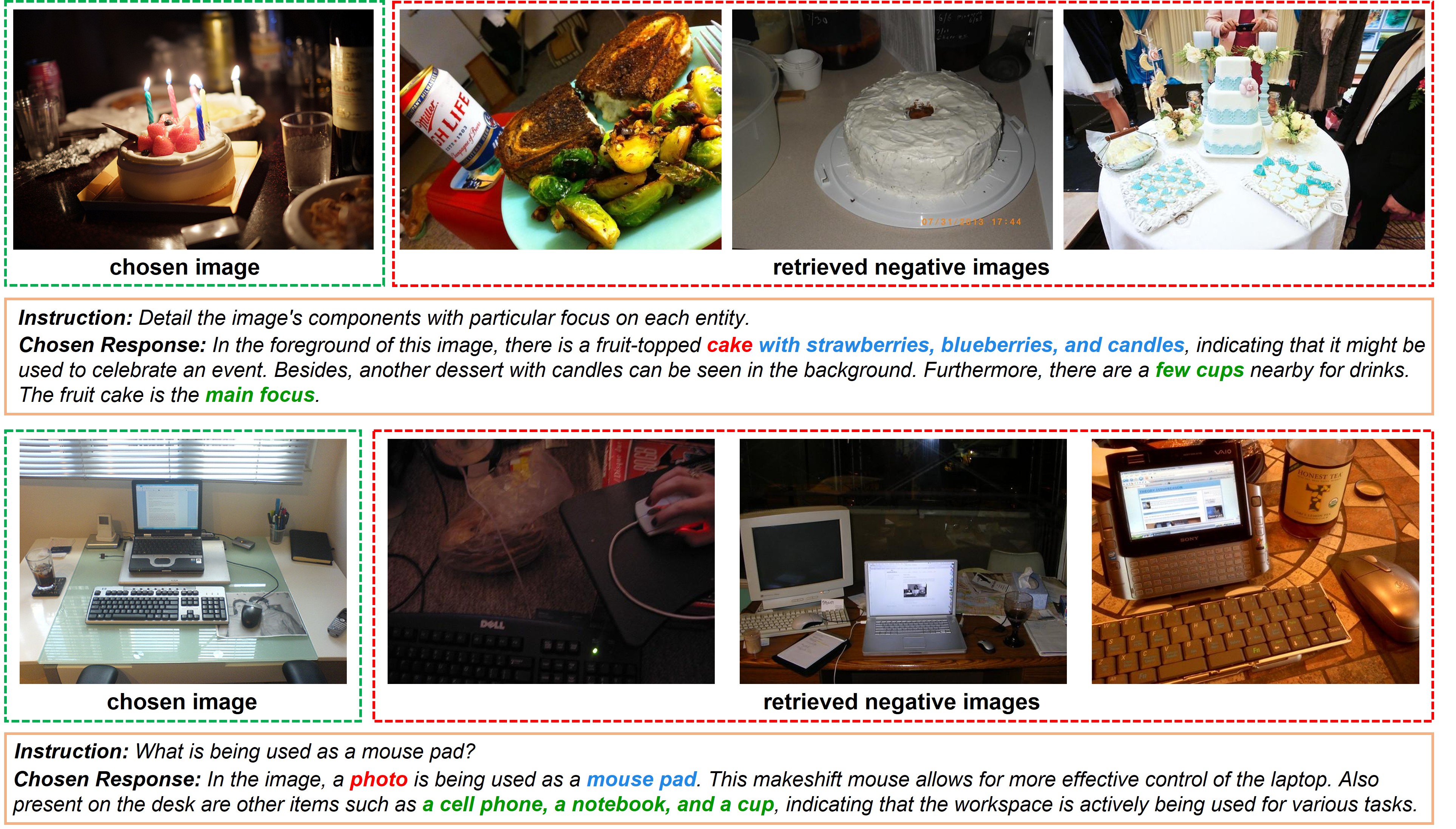}
    \vspace{-1mm}
    \caption{
    Negative image retrieval using our \ourmodel method. Each row shows a chosen image and three negatives; highlighted phrases in \textbf{\textcolor{red}{red}}, \textbf{\textcolor{blue}{blue}} and \textbf{\textcolor{green!70!black}{green}} mark mismatches with negatives.}
    \label{fig:sampling-examples}
    \vspace{-7mm}
\end{figure}

\subsection{Effectiveness of Importance Sampling}
We analyze the impact of our importance sampling strategy using t-SNE visualizations of high-quality negative images, shown in Figure~\ref{fig:tsne-sampling}. The left plot displays negatives selected by our SAE-guided strategy, while the middle shows randomly sampled ones. Importance-sampled negatives are well-dispersed across the embedding space, indicating high semantic diversity, whereas random negatives cluster tightly, indicating redundancy and limited coverage. This distribution reveals that our method captures heterogeneous error modes rather than collapsing onto a narrow, repetitive subset of examples, thereby providing more informative supervision.

As illustrated in Figure~\ref{fig:sampling-examples}, our selected negatives for the fruit cake image exhibit clear semantic deviations: one introduces different types of cakes with distinct decorations, while others include unrelated food items such as sandwiches, or scenes where the cake is present but not the main visual focus, thereby weakening semantic alignment. Similarly, in the workspace scene, negative examples capture meaningful variations such as differences in desk arrangement and surrounding objects—including notebooks and cups—each impacting multimodal alignment differently.
These examples highlight how our sampling method uncovers diverse error modes, encouraging the model to learn more robust visual distinctions, strengthening training signals, and improving generalization beyond narrow, one-dimensional deviations.

\begin{table*}[tbp]
  \centering
  \renewcommand{\arraystretch}{1.2}
  \setlength{\tabcolsep}{4pt} 
  \resizebox{\linewidth}{!}{
  \begin{tabular}{@{} 
      l   
      l   
      cc  
      ccc 
      c   
      cc  
      c   
    @{}}
    \toprule
    \textbf{Model} & \textbf{Benchmark}
      & \multicolumn{2}{c}{\textbf{MMHalBench}}
      & \multicolumn{3}{c}{\textbf{HallusionBench}}
      & \textbf{POPE}
      & \multicolumn{2}{c}{\textbf{WildVision}}
      & \textbf{MMVP} \\
    \cmidrule(lr){3-4}\cmidrule(lr){5-7}\cmidrule(lr){8-8}\cmidrule(lr){9-10}\cmidrule(lr){11-11}
    & 
      & \textbf{Score} ($\uparrow$)
      & \textbf{HalRate} ($\downarrow$)
      & \textbf{aA} ($\uparrow$)
      & \textbf{fA} ($\uparrow$)
      & \textbf{qA} ($\uparrow$)
      & \textbf{Acc.} ($\uparrow$)
      & \textbf{Reward} ($\uparrow$)
      & \textbf{WinRate} ($\uparrow$)
      & \textbf{Acc.} ($\uparrow$)\\
    \midrule
    \multirow{5}{*}{\textbf{Qwen2.5-7B}}
     & mdpo 
      & 5.01 & 14.89 & 67.40 & 41.33 & 42.20 & 87.02 & 28.5 & 66.2  & 76.33 \\
      & diffusion
      & \textbf{5.12}     & 12.50     & 69.50 & 43.64 & 43.95 & 87.52 & 30.7 & 66.4 & 78.00  \\
    & crop+diffusion
      & 4.92    & 13.54   & 70.35  & 43.35  & 44.61 & 87.35 & 30.7  &  66.4 & 78.33    \\
    & similarity & 5.00   & 12.50 & 69.82  & 42.77 & 44.17 & 87.24 & 30.1 & 66.2 & 77.67 \\
    & \textbf{\ourmodel}
       & 5.05  & \textbf{11.46}  & \textbf{71.24} & \textbf{43.77} & \textbf{45.61} & \textbf{88.66} & \textbf{32.4}  & \textbf{68.4}  & \textbf{79.00}  \\
    \midrule
    \multirow{5}{*}{\textbf{Qwen2.5-3B}}
     & mDPO   & 4.47  & 21.28 & 62.88 & 35.84 & 37.14 & 87.65 & 7.2  & 50.8  & 69.33 \\
      & diffusion
      & 4.56 & 14.58 & 65.19 & \textbf{39.31} & 37.14 & 87.35 & 8.0 & \textbf{52.4} & 71.33  \\
    & crop+diffusion
     & 4.39   & 17.70   & 64.03 & 38.44  & 36.48 & 87.20 & 5.3  & 49.8 & 71.33    \\
    & similarity & 4.20 & 19.79 & 65.08 & 39.01 & 36.92 & 87.62 & 4.7 & 49.4 & 70.30 \\
    & \textbf{\ourmodel}
      & \textbf{4.61}  & \textbf{13.54}  & \textbf{65.51}  & 38.44  & \textbf{38.02} & \textbf{87.77} & \textbf{8.6}  & \textbf{52.4}  & \textbf{72.00}      \\
    \bottomrule
  \end{tabular}
  }
  \caption{
Comparison of different negative sampling strategies across five benchmarks. All variants share the same model and loss design, differing only in how negative images are constructed.
}
\label{tab:diffusion}
\vspace{-3mm}
\end{table*}

\subsection{Comparison of Negative Sampling Strategies}
We evaluate five negative sampling strategies that share the same model architecture and loss formulation but differ in how negative images are constructed:
(1) mDPO, which relies on a single diffusion-generated negative;
(2) diffusion, which combines one diffusion negative with two negatives selected by our method;
(3) crop+diffusion, which mixes one cropped, one diffusion, and one \ourmodel-selected negative;
(4) similarity, where three negatives are retrieved based on similarity to the positive image; and
(5) our model.
As shown in Table~\ref{tab:diffusion}, mDPO yields the weakest performance, highlighting the limitations of single-negative supervision. 
Among multi-negative variants, the diffusion strategy outperforms crop+diffusion, suggesting that a higher proportion of semantically diverse negatives from our method improves supervision quality. 
The similarity variant performs worse than diffusion, underscoring that naive retrieval of visually similar negatives does not provide the challenging guidance needed. 
In contrast, \ourmodel consistently achieves the best scores across all benchmarks, validating the effectiveness of structured multi-negative selection via SAE-guided importance sampling. 
Additional results are reported in Table~\ref{tab:llava} in the appendix.

Beyond the choice of sampling strategy, we further explore the number of negatives required. Figure~\ref{fig:tsne-sampling} (right) shows performance across HallusionBench, MMVP, and MMHalBench with varying numbers of negatives. One or two negatives are insufficient to provide high-quality supervision. Performance peaks at three negatives, while increasing beyond this offers no further benefit. For HallusionBench, adding five negatives even reduces performance, likely due to noise introduced by redundant or low-quality samples. Together, these results demonstrate that three carefully chosen negatives strike the best balance between informativeness and robustness.

\subsection{Hallucination Reduction with \ourmodel.}
Figure~\ref{fig:hallucination_examples} in the Appendix presents qualitative comparisons highlighting the impact of multi-negative supervision.
Baseline methods such as DPO and CHiP frequently introduce hallucinated details (e.g., incorrect objects, colors, or spatial relations), while \ourmodel generates more faithful and grounded descriptions. 
For instance, in the first example, only \ourmodel correctly identifies that the image lacks sand and accurately describes the bird. 
These results illustrate that incorporating diverse negatives enables the model to better distinguish relevant from spurious cues, improving factual accuracy in vision-language alignment.

\section{Conclusion}


We present \ourmodel, a novel framework that introduces multi-negative, semantically diverse supervision into multimodal Direct Preference Optimization. By leveraging CLIP-based embeddings and a sparse autoencoder, our method efficiently selects image-side negatives that vary across multiple semantic facets and reflect diverse failure modes. These negatives are integrated into a Plackett-Luce-style ranking objective with importance sampling, enabling the model to learn from richer and more structured supervision. The method remains efficient and scalable for real-world multimodal applications. Extensive experiments across five benchmarks demonstrate that \ourmodel consistently outperforms strong baselines in hallucination reduction and visual grounding. While our evaluations rely on GPT-based scoring—which may introduce bias or inconsistency when assessing fine-grained alignment—our findings validate the effectiveness of semantic-aware, multi-negative sampling for robust multimodal alignment and open up promising directions for scalable and interpretable preference-based learning.



\bibliography{iclr2026_conference}
\bibliographystyle{iclr2026_conference}

\newpage

\appendix
\section{Appendix}


\subsection{Effect of $\beta$ on Hallucination and Quality}

\begin{figure*}[ht]
    \centering
    \includegraphics[width=1.0\linewidth]{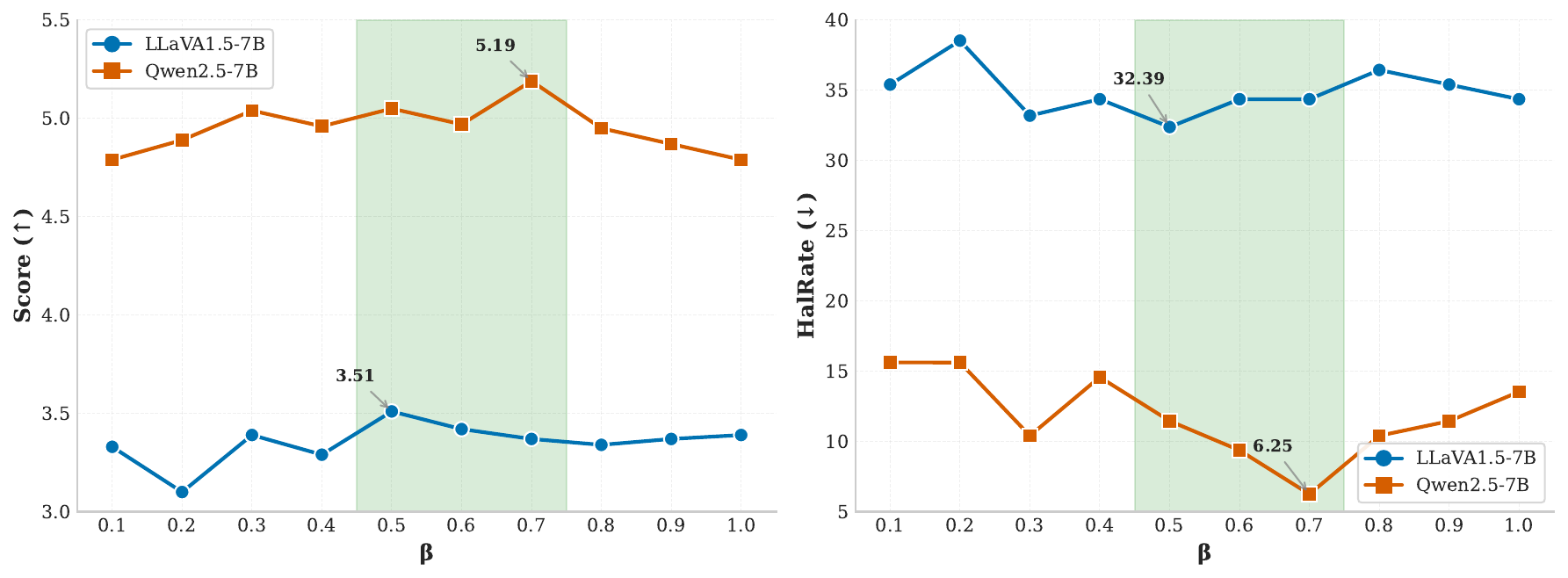}
    \vspace{-3mm}
  \caption{Performance comparison of Score and Hallucination Rate across different $\beta$ values for on MMHalBench.}
    \label{fig:beta_value}
    \vspace{-3mm}
\end{figure*}

Figure~\ref{fig:beta_value} illustrates the impact of different $\beta$ values on both response quality (left) and hallucination rate (right) on MMHalBench. We observe that the model performs best when $\beta$ lies in the range of 0.45 to 0.75, achieving a good balance between response quality and hallucination suppression. To ensure both accuracy and stability, we set $\beta=0.5$ as the default value in all main experiments.

\subsection{Examples of Hallucination Reduction with MISP-DPO}
To illustrate the impact of our approach, Figure~\ref{fig:hallucination_examples} presents qualitative comparisons across three representative prompts, highlighting the improvements brought by \ourmodel over baselines including LLaVA, DPO, and CHiP. \ourmodel demonstrates a stronger ability to avoid hallucinations and produce factually accurate descriptions grounded in visual evidence. In the first example, it correctly identifies the absence of sand and avoids misidentifying the bird species. In the second case, it faithfully describes the structure and positioning of the gloves despite occlusion. In the final example, it provides a precise spatial interpretation of the two watches without fabricating brand-specific details. These results suggest that our multi-negative supervision strategy improves the model’s sensitivity to fine-grained semantic cues and its ability to reject spurious correlations and hallucinated attributes, leading to more factually consistent vision-language generation.

\

\subsection{Negative Sampling Strategies on LLaVA-1.5-7B model}
Table~\ref{tab:llava} shows that LLaVA-1.5-7B follows the same trend as Qwen2.5. The single-negative baseline (mDPO) performs the weakest, while adding more negatives through diffusion or crop+diffusion gives only moderate improvements.

In contrast, \ourmodel consistently achieves the best results, notably lowering hallucination rate and improving accuracy across HallusionBench, POPE, WildVision, and MMVP. This confirms that our SAE-guided multi-negative sampling generalizes across model families, providing stronger guidance even for smaller-scale vision–language models.

\begin{table*}[ht]
  \centering
  \renewcommand{\arraystretch}{1.2}
  \setlength{\tabcolsep}{4pt} 
  \resizebox{\linewidth}{!}{
  \begin{tabular}{@{} 
      l   
      l   
      cc  
      ccc 
      c   
      cc  
      c   
    @{}}
    \toprule
    \textbf{Model} & \textbf{Benchmark}
      & \multicolumn{2}{c}{\textbf{MMHalBench}}
      & \multicolumn{3}{c}{\textbf{HallusionBench}}
      & \textbf{POPE}
      & \multicolumn{2}{c}{\textbf{WildVision}}
      & \textbf{MMVP} \\
    \cmidrule(lr){3-4}\cmidrule(lr){5-7}\cmidrule(lr){8-8}\cmidrule(lr){9-10}\cmidrule(lr){11-11}
    & 
      & \textbf{Score} ($\uparrow$)
      & \textbf{HalRate} ($\downarrow$)
      & \textbf{aA} ($\uparrow$)
      & \textbf{fA} ($\uparrow$)
      & \textbf{qA} ($\uparrow$)
      & \textbf{Acc.} ($\uparrow$)
      & \textbf{Reward} ($\uparrow$)
      & \textbf{WinRate} ($\uparrow$)
      & \textbf{Acc.} ($\uparrow$)\\
    \midrule
    \multirow{3}{*}{\textbf{llava-1.5-7b}}
     & mDPO    & 2.99 & 49.81 & 47.32 & 20.52 & 13.19 & 83.25 & -62.1 & 14.6 & 58.33\\
      & diffusion
      & 3.49 & 33.33 & 52.57 & 21.38 & 18.02 & 83.25 & –53.0 & 18.8  & 61.33 \\
    & crop+diffusion
      & 3.44 & 35.42 & 53.63 & 22.54 & 19.78 & 83.80 & -50.5   & 20.0 & 61.00  \\
    & \textbf{\ourmodel}
      & \textbf{3.51}  & \textbf{32.29} & \textbf{57.52}   & \textbf{25.43}   & \textbf{24.83} & \textbf{83.94} & \textbf{-46.4} & \textbf{20.6} & \textbf{63.00}  \\
    \bottomrule
  \end{tabular}
  }
  \caption{llava performance}
\label{tab:llava}
\end{table*}

\begin{figure}[ht]
\centering

{\fontfamily{cmbr}\selectfont \scriptsize

\setlength{\tabcolsep}{2pt}
\renewcommand{\arraystretch}{1.05}
\arrayrulecolor{gray!40}

\begin{tcolorbox}[
  colframe=black!50,
  colback=white,
  boxrule=0.4pt,
  arc=2pt,
  width=1.0\linewidth,
  boxsep=1pt,
  top=2pt, bottom=2pt
]
\begin{minipage}[t]{0.19\linewidth}
\vspace{-2pt}
\raisebox{0pt}[0pt][0pt]{\includegraphics[width=\dimexpr\linewidth-1.5pt\relax]{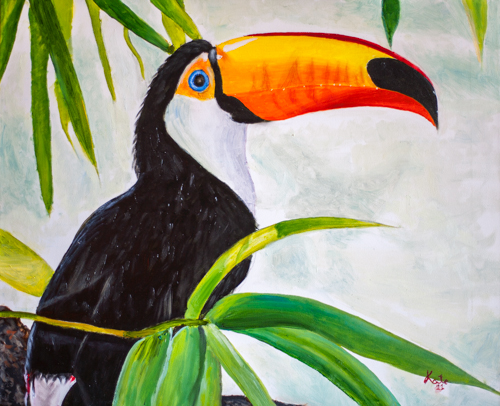}}
\vspace{-4pt}
\begin{tcolorbox}[
  colback=purple!8,
  colframe=purple!10,
  boxrule=0.3pt,
  sharp corners=all,
  boxsep=2pt,
  left=2pt, right=2pt,
  top=1pt, bottom=1pt,
  width=\linewidth
]
\raggedright
\textbf{What is a vicuna standing in the sand looking at?}
\end{tcolorbox}
\end{minipage}
\hfill
\begin{minipage}{0.78\linewidth}
\begin{tabular}{@{}p{0.13\linewidth}@{\hspace{4pt}}p{0.82\linewidth}@{}}
\textbf{LLaVA} & A \textcolor{red!75!black}{vicuna standing in the sand} is \textcolor{red!75!black}{looking at a painting of a bird}.\\
\cmidrule(lr){1-2}
\textbf{DPO} & A \textcolor{red!75!black}{vicuna standing in the sand} is not looking at anything. In the image, \textcolor{blue!70!black}{a black and white bird with a yellow and red beak is standing in front of a green leafy plant}. The bird is not looking at the plant, but rather at the viewer. \\
\cmidrule(lr){1-2}
\textbf{CHiP} & A \textcolor{red!75!black}{vicuna standing in the sand} is \textcolor{red!75!black}{looking at a painting of a bird}. \\
\cmidrule(lr){1-2}
\textbf{MISP-DPO (Ours)} & In the image, there is \textcolor{blue!70!black}{no sand} or any other background visible. The painting features a \textcolor{blue!70!black}{colorful bird with a large beak}. It is a black and white bird with a yellow beak, \textcolor{blue!70!black}{not a vicuna}. The bird is surrounded by green and black brush, and the painting is described as a painting of a bird with a yellow beak. The bird is not standing in the sand, and the image \textcolor{blue!70!black}{does not provide any context or information about the bird looking at something}. \\
\end{tabular}
\end{minipage}
\end{tcolorbox}

\vspace{1pt}

\begin{tcolorbox}[
  colframe=black!50,
  colback=white,
  boxrule=0.4pt,
  arc=2pt,
  width=1.0\linewidth,
  boxsep=1pt,
  top=2pt, bottom=2pt
]
\begin{minipage}[t]{0.19\linewidth}
\vspace{-2pt}
\raisebox{0pt}[0pt][0pt]{\includegraphics[width=\dimexpr\linewidth-1.5pt\relax]{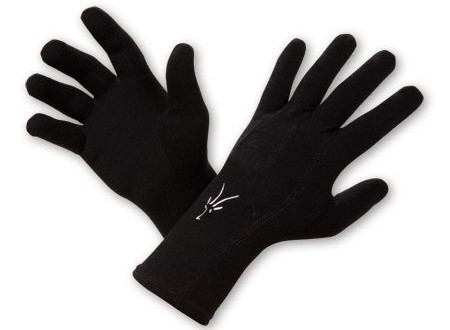}}
\vspace{-4pt}
\begin{tcolorbox}[
  colback=purple!8,
  colframe=purple!10,
  boxrule=0.3pt,
  sharp corners=all,
  boxsep=2pt,
  left=2pt, right=2pt,
  top=1pt, bottom=1pt,
  width=\linewidth
]
\raggedright
\textbf{Describe the hands\\wearing the gloves.}
\end{tcolorbox}
\end{minipage}
\hfill
\begin{minipage}{0.78\linewidth}
\begin{tabular}{@{}p{0.13\linewidth}@{\hspace{4pt}}p{0.82\linewidth}@{}}
\textbf{LLaVA} & The \textcolor{red!75!black}{hands wearing the gloves} are large, and the gloves are \textcolor{blue!70!black}{black} in color. \\
\cmidrule(lr){1-2}
\textbf{DPO} & The \textcolor{red!75!black}{hands wearing the gloves} are black and have \textcolor{red!75!black}{a logo on the palm}. The gloves are made of a \textcolor{blue!70!black}{black material}, and the hands are positioned in a way that they are \textcolor{red!75!black}{facing upwards}. \\
\cmidrule(lr){1-2}
\textbf{CHiP} & In the image, the \textcolor{red!75!black}{hands wearing the gloves} are \textcolor{blue!70!black}{black} in color. \\
\cmidrule(lr){1-2}
\textbf{MISP-DPO (Ours)} & The image shows two \textcolor{blue!70!black}{black gloves}, one of which has \textcolor{blue!70!black}{a logo on the back}. They are positioned in such a way that they appear to be two hands. The gloves are black and have a close-up view, making it difficult to discern any other details. The gloves are \textcolor{blue!70!black}{not attached to any hands}, but they are positioned in a way that they appear to be hands. \\
\end{tabular}
\end{minipage}
\end{tcolorbox}

\vspace{1pt}

\begin{tcolorbox}[
  colframe=black!50,
  colback=white,
  boxrule=0.4pt,
  arc=2pt,
  width=1.0\linewidth,
  boxsep=1pt,
  top=2pt, bottom=2pt
]
\begin{minipage}[t]{0.19\linewidth}
\vspace{-2pt}
\raisebox{0pt}[0pt][0pt]{\includegraphics[width=\dimexpr\linewidth-1.5pt\relax]{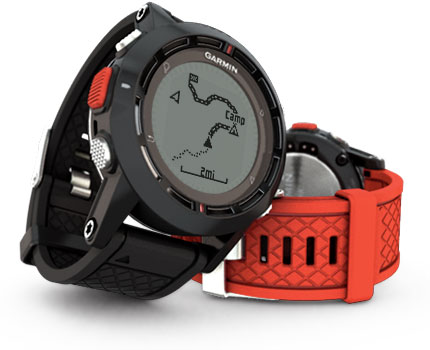}}
\vspace{-4pt}
\begin{tcolorbox}[
  colback=purple!8,
  colframe=purple!10,
  boxrule=0.3pt,
  sharp corners=all,
  boxsep=2pt,
  left=2pt, right=2pt,
  top=1pt, bottom=1pt,
  width=\linewidth
]
\raggedright
\textbf{Describe how the two watches are placed.}
\end{tcolorbox}
\end{minipage}
\hfill
\begin{minipage}{0.78\linewidth}
\begin{tabular}{@{}p{0.13\linewidth}@{\hspace{4pt}}p{0.82\linewidth}@{}}
\textbf{LLaVA} &   In the image, there are \textcolor{blue!70!black}{two watches placed next to each other}. One watch is positioned on top of the other, with the \textcolor{red!75!black}{top watch being a G-Shock watch and the bottom watch being a Garmin watch}. The two watches are placed side by side, \textcolor{blue!70!black}{showcasing their differences in design and features}.\\
\cmidrule(lr){1-2}
\textbf{DPO} & The \textcolor{red!75!black}{watch with the red band is positioned on top} of the watch with the black band.The \textcolor{red!75!black}{watch with the red band is on the left side of the image}, while the watch with the black band is on the left side of the image as well. \\
\cmidrule(lr){1-2}
\textbf{CHiP} & In the image, one watch is in the \textcolor{red!75!black}{foreground}, and the other is in the \textcolor{red!75!black}{background}. The foreground one is a \textcolor{red!75!black}{close-up} while the other is \textcolor{red!75!black}{blurry}. \\
\cmidrule(lr){1-2}
\textbf{MISP-DPO (Ours)} & The two watches are \textcolor{blue!70!black}{placed side by side}, with the \textcolor{blue!70!black}{Garmin watch on the left} and the other watch on the right. The \textcolor{blue!70!black}{Garmin watch has a screen} displaying a \textcolor{blue!70!black}{map}, while the \textcolor{blue!70!black}{other watch has a red band}. The two watches are positioned in such a way that \textcolor{blue!70!black}{the Garmin watch's screen is visible, and the other watch's red band is also visible}.\\
\end{tabular}
\end{minipage}
\end{tcolorbox}
}

\vspace{-1mm}
\caption{Comparison of responses from LLaVA model training through different methods: pretrained, DPO, CHiP, and MISP-DPO. \textcolor{blue!70!black}{Blue text} indicates faithful descriptions; \textcolor{red!75!black}{red text} marks hallucinated or unsupported content.}
\label{fig:hallucination_examples}
\end{figure}

\end{document}